%% file: emnlp2018.tex
\documentclass[11pt,a4paper]{article}
\usepackage[hyperref]{emnlp2018}
\usepackage{times}
\usepackage{latexsym}
\usepackage{booktabs}
\usepackage{microtype}
\usepackage[inline]{enumitem}

\usepackage{url}
\usepackage{graphicx}

\aclfinalcopy % Uncomment this line for the final submission

\title{Not Just Depressed: Bipolar Disorder Prediction on Reddit}

\author{Ivan Sekuli{\'{c}} \quad Matej Gjurkovi{\'{c}} \quad Jan {\v{S}}najder\\
Text Analysis and Knowledge Engineering Lab\\
Faculty of Electrical Engineering and Computing, University of Zagreb\\
Unska 3, 10000 Zagreb, Croatia \\
\tt \{ivan.sekulic,matej.gjurkovic,jan.snajder\}@fer.hr
}

\date{}

\begin{document}
\maketitle
\begin{abstract}
Bipolar disorder, an illness characterized by manic and depressive episodes,
affects more than 60 million people worldwide. We present a preliminary study
on bipolar disorder prediction from user-generated text on Reddit, which relies
on users' self-reported labels. Our benchmark classifiers for bipolar disorder
prediction outperform the baselines and reach accuracy and F1-scores of above
86\%. Feature analysis shows interesting differences in language use between
users with bipolar disorders and the control group, including differences in
the use of emotion-expressive words.
\end{abstract}

\input{introduction}

\input{related}

\input{dataset}

\input{prediction}

\input{feature}

\input{emotions}

\input{discussion}

\bibliography{emnlp2018} 
\bibliographystyle{acl_natbib_nourl}

\end{document}

%% file: introduction.tex
\section{Introduction}

% http://www.who.int/news-room/fact-sheets/detail/mental-disorders
% . In low- and middle-income countries, between 76% and 85%

World Health Organization's 2017 and
\citet{wykes2015mental} report that up to $27\%$ of adult population in Europe suffer or
have suffered from some kind of mental disorder.  Unfortunately, as much as
35--50\% of those affected go undiagnosed and receive no treatment for their
illness.  To counter that, the WHO's Mental Health Action Plan's
\citep{saxena2013world} lists as one of its main
objectives the gathering of information and evidence on mental conditions.  At the same time, analysis of texts
produced by authors affected by mental disorders is attracting increased
attention in the natural language processing community.  The research is geared
toward a deeper understanding of mental health and the development of models
for early detection of various mental disorders, especially on social networks.

In this paper we focus on bipolar disorder, a complex psychiatric disorder manifested by uncontrolled changes in mood and energy levels.
Bipolar disorder is characterized by manic episodes, during which people feel abnormally elevated and energized, and depression episodes, manifested in decreased activity levels and a feeling of hopelessness. % \citet{anderson2012bipolar} mozda i ne citirati jer realno nije od njega definicija neg ovak?
The two phases are recurrent and differ in intensity and duration, greatly affecting the person's capacity to carry out daily tasks.
%Severe manic episodes are typical for bipolar 1 disorder, causing trouble sleeping, reckless behaviour etc., while less sever episodes (hypomania) are characteristic for bipolar 2 disorder.
Bipolar disorder affects more than 60 million people, or almost 1\% of the world population \citep{anderson2012bipolar}.
%although research suggests that the lifetime prevalence may be as high as 6.4\% if one takes into account the subthreshold cases \citep{judd2003prevalence}. 
% The prevalence is equal in men and women,/
The suicide rate in patients diagnosed with bipolar disorder is more than 6\% \citep{nordentoft2011absolute}. There is thus a clear need for the development of systems capable of early detection of this illness.

%This kind of analysis requires a rich source of user-generated text, which can be found online, specifically on forums, discussion platforms and popular social media websites where people often come to talk, seek and offer support regarding their mental illness.

As a first step toward that goal, in this paper we present a preliminary study
on bipolar disorder prediction based on user-generated texts on social media.
The main problem in detecting mental disorders from user-generated text is the
lack of labeled datasets. We follow the recent strand of research
\citep{gkotsis2016language, de2016discovering, shen2017detecting,
gjurkovic2018reddit} and use Reddit as a rich and diverse source of high-volume
data with self-reported labels. Our study consists of three parts. First, we test benchmark models for predicting Reddit users 
with bipolar disorder. Second, we carry out a
feature analysis to determine which psycholinguistic features are good
predictors of the disorder. Lastly, acknowledging that emotional swings are the
main symptom of the disorder, we analyze the emotion-expressive textual
features in bipolar disorder users and the non-bipolar control group of users.

%% file: related.tex
\section{Related Work}

Psychologist have long studied the language use in patients with mental
disorders, including schizophrenia \citep{taylor1994patterns}, suicidal
tendencies \citep{thomas1985words}, and depression
\citep{schnurr1992comparison}. Lately, computer-based analysis with LIWC (Linguistic Inquiry and Word Count) \citep{pennebaker2001linguistic}
resource was used to extract features for various studies regarding mental health \cite{pennebaker1999linguistic}.
%Computer-based analysis begins with the General
%Inquirer \citep{stone1966general}, which uses dictionaries of words grouped
%into several categories, an approach later made popular by the LIWC (Linguistic
%Inquiry and Word Count) \citep{pennebaker2001linguistic} resource. LIWC groups
%words into 90 categories, which were used as features for various studies
%regarding mental health \cite{pennebaker1999linguistic}. 
For example,
\citet{stirman2001word} found the increased use of the first-person singular 
pronouns (\emph{I, me, my}) in poems to be a good predictor of suicidal 
behavior, while
\citet{rude2004language} detected an excessive use of the pronoun \emph{I} in
essays of depressed psychology students.  In a recent study, however,
\citet{tackman2018depression} suggest that first-person singular pronouns may
be better viewed as a marker of general distress or negative emotionality
rather than as a specific marker of depression.

%Also, the correlation between positive emotions and the pronoun \emph{you} is significant, which can be related to the origin of the dataset. % or can it?

%However, they do not report if the results are statistically significant.
%They compare their text analysis tool, similar to LIWC, with previous studies on emotional writing and casual conversation. % referenca????
% nista nema u clanku...

A number of studies looked into the use of emotion-expressive words.
\citet{rude2004language} found that currently depressed students used more
negative emotion words than never-depressed students.
\citet{halder2017modeling} tracked linguistic changes of social network users
over time to understand the progression of their emotional status.
%, including
%the prediction of future emotional status from user's past posts.
\citet{kramer2004text} 
%focused on online bipolar disorder users and 
found that
conversations in bipolar support chat rooms contained more positively valence
words and slightly more negatively valenced emotions than casual conversations.
%\maybe{Furthermore, they found a significant positive
%correlation between the use of words related to bipolar disorder and both negative
%emotions and the pronoun \emph{I}.}

%\todo{more RW on mental
%disorders+emotions?}

%emotions...

%Self-referential language using first-person singular pronouns may therefore be better construed as a ling
%uistic marker of general distress proneness or negative emotionality rather than as a specific marker of depression.

Much recent work has leveraged social media as a source of user-generated 
text for mental health profiling \citep{park2012depressive}.  Most studies used
Twitter data; e.g., \citet{de2013social} predicted depression in Twitter users,
while CLPsych 2015 shared task \citep{coppersmith2015clpsych} addressed
depression and post-traumatic stress disorder (PTSD). 
Bipolar disorder on Twitter is usually classified alongside other disorders. E.g., \citet{coppersmith2014quantifying,coppersmith2015adhd} achieved a precision of 0.64 at 10\% false alarms, while
\citet{benton2017multi} used multi-task learning and achieved an AUC-score of 0.752.
%\citet{beller2014ma} and
%\citet{coppersmith2014quantifying} made use of self-reporting signals in users'
%tweets to build a dataset labeled with various disorders, achieving precision of 0.64 at 10\% false alarms for bipolar disorder.
%including bipolar
%disorder. In particular, on bipolar disorder prediction their model achieves a
%precision (diagnosed, correctly labeled) of 0.64 at 10\% false alarms (control,
%labeled as diagnosed). 
%A similar result was reported by
%\citep{coppersmith2015adhd}, who addressed ten different mental disorders.
%\citet{benton2017multi} used multi-task learning to built a
%joint model for multiple mental disorders, achieving an AUC score of 0.752 for
%bipolar disorder.

%The volume of the datasets expanded with the hunt for self-reported mental health issues on social media, by searching text like "I was diagnosed with depression" in posts.
%By employing this method, proposed by \citet{beller2014ma}, \citet{coppersmith2014quantifying} generate a large-scale dataset, consisted of $441$ depressed Twitter users, $244$ users with PTSD, $159$ users with seasonal affective disorder (SAD) and $394$ bipolar users.
%On a binary classification task, bipolar users vs.~control group, they achieve precision of $0.64$ at $10\%$ false alarms and $0.82$ at $20\%$.
%Similar precision for predicting users with bipolar disorder is reported in their other work \citep{coppersmith2015adhd}, where they try to tackle ten different mental disorders.

Reddit has only recently been used as a source for the analysis of mental
disorders. \citet{gkotsis2016language} analyzed the language in different
subreddits related to mental health, and showed that linguistic features such
as vocabulary use and sentence complexity vary across different subreddits.
%whether a comment belongs to
%subreddit related to mental health.
%, and that negative sentiment prevails inthose subreddits. 
\citet{de2016discovering} explored the methods for automatic
detection of individuals which could transit from mental health discourse to
suicidal ideas.  \citet{shen2017detecting} used topic modeling, LIWC, and language models to predict whether a Reddit post is
related to anxiety.  To our
knowledge, there is no previous study on the analysis of bipolar disorder of
Reddit users.

%, alongside depression and PTSD, deals with seasonal affective disorder (SAD) and bipolar disorder in Twitter.

%% file: dataset.tex
\section{Dataset}

Reddit is one of the largest social media sites in the world, with more than 85
million unique visitors per
month.\footnote{https://www.alexa.com/siteinfo/reddit.com} Reddit is suitable
for our study not only because of its vast volume, but also because it offers
user anonimity and covers a wide range of topics. Registered users can
anonymously discuss various topics on more than 1 million subpages, called
``subreddits''.  A considerable number of subreddits is dedicated to mental
health in general, and to bipolar disorder in particular. All comments between
2005 and 2018 (more than 3 billion) are available as a Reddit dump database via
Google Big Query, which we used to obtain the data.

\paragraph{Bipolar disorder users.}

To obtain a sample of users with bipolar disorder, we first retrieved all
subreddits related to the disorder, i.e., \emph{bipolar}, \emph{bipolar2},
\emph{BipolarReddit}, \emph{BipolarSOs}, \emph{bipolarart}, as well as the more generic \emph{mentalhealth} subreddit.
Then, following \citet{beller2014ma} and \citet{coppersmith2014quantifying}, we
looked for self-reported bipolar users by searching in the user's comments for the string 
\emph{``I am diagnosed with bipolar''} and its paraphrased versions. In addition, following
\citet{gjurkovic2018reddit}, we inspect users' \emph{flairs} -- short
descriptive texts that the users can set for certain subreddits to appear next
to their names. While a flair is not mandatory, we found that many users with bipolar disorder do use flairs on mental health subreddits to indicate their disorder.

%Thus, we query the database for users that express their illness in \emph{flairs}, searching for words like \emph{Bipolar} or \emph{BP1} (indicating the bipolar type 1).

The acquisition procedure yielded a set of 4,619 unique users with self-reported
bipolar disorder.  The users generated around 5 million comments, totaling more
than 163 million tokens.  To get an estimate of labeling quality, we randomly
sampled 250 users and inspected their labels and text. As we found no false
positives (i.e., all 250 users report on being diagnosed a bipolar disorder),
we gauge that the dataset is of high precision. 
The true precision of the dataset depends, of course, on the veracity of
the self-reported diagnosis.

To make the subsequent analysis reliable and unbiased, we decided to
additionally prune the dataset as follows.  To mitigate the topic bias, we
removed all comments by bipolar disorder users on bipolar subreddits, as well
as on the general mental health subreddit.  Additionally, any comment on any
subreddit that mentions the words \emph{bipolar} or \emph{BP} (case insensitive)
was also excluded. Finally, to increase the reliability,
we retained in our dataset only the users who, after pruning, have at least
1000 word remaining. The final number of users in our dataset is 3,488.

\paragraph{Control group.}

The control group was sampled from the general Reddit community, serving as
a representative of the mentally healthy population. To ensure that
the topics discussed by the control group match those of 
bipolar disorder users, we sampled users that post in subreddits often
visited by bipolar disorder users (i.e., subreddits where posting frequency of
bipolar disorder users was above the average). To also ensure
that the control group is representative of the mentally healthy Reddit
population, we removed all users with more than 1000 words on mental health
related subreddits. As before, we only retained users that had more than 1000
words in all of their comments. The final number of users in the
control group is 3,931, which is close to the number of bipolar users, with the purpose of having a 
balanced dataset. The total number of comments is about 20 million,
which is four times more than for the bipolar disorder users.

%It is composed of
%randomly chosen $4200$ authors that post in subreddits often visited by bipolar
%users.  Big Query returned a list of authors that post in at least one of the
%ten subreddits of different topics, taken from the ones bipolar users post more
%frequently than average.  As before, the user had to have more than $1000$
%words in total in all of their comments.  The $4200$ users had more than $20$
%million comments in total, approximately $4$ times more than the bipolar group.
%To maintain our control group as clean as possible, we remove users with more
%than $1000$ words on mental health related subreddits, as we found some
%reporting to suffer from various mental disorders.

\paragraph{Topic categories.}

Topic of discussion may affect the language use, including the stylometric
variables \cite{mikros2007investigating}, which means that topic distribution
may act as a confounder in our analysis. To minimize this effect, we split the
dataset into nine topic categories, each consisting of a handful of subreddits
on a similar topic. Table \ref{tbl:categories} shows the breakdown of the
number of unique users from both groups across topic categories.
\emph{AskReddit} is the biggest subreddit and not bound to any particular topic; in this category, we also add other subreddits covering a wide range of
topics, such as \emph{CasualConversation} and \emph{Showerthoughts}.
%Other categories are self-explanatory.
%, e.g., the category \emph{Sex and
%relationships} covers subreddits such as \emph{relationship\_advice},
%\emph{TwoXChromosomes}, and \emph{asktransgender}. 
To be included in a
category, the user must have had at least 1000 words on subreddits from that
category.

%In this way, we can also analyze the differences in language across different topics.
%The topics are: \emph{Animals}, \emph{AskReddit}, \emph{Gaming}, \emph{Jobs and finance}, \emph{Movies/music/books}, \emph{Politics}, \emph{Religion}, \emph{Sex and relationships}, \emph{Sports}.

%The distribution of bipolar users and users from the control group varies through categories, as shown in table \ref{tbl:categories}.
%As mentioned, control users have $4$ times more comments than bipolar, so the distribution doesn't come as a surprise.
%The analysis of psycholinguistic features, as well as the classification results, are reported for the whole dataset, and for each of the categories.

\begin{table}
\centering
{\small
\begin{tabular}{lrr}
\toprule
Category            & \#~bipolar   & \#~control   \\
\midrule
Animals               & 397  & 898  \\
AskReddit             & 1797 & 2767 \\
Gaming                & 489  & 1501 \\
Jobs and finance      & 293  & 586  \\
Movies/music/books    & 502  & 1606 \\
Politics              & 332  & 2445 \\
Religion              & 264  & 700  \\
Sex and relationships & 948  & 1000 \\
Sports                & 156  & 785 \\
\midrule
All                   & 3488 & 3931 \\
\bottomrule
\end{tabular}}
\caption{\label{tbl:categories} The number of unique bipolar disorder and control group users broken down by topic categories}
\end{table}

%% file: prediction.tex
\section{Bipolar Disorder Prediction}

\paragraph{Feature extraction.}

For each user, we extracted three kinds of features: (1) psycholinguistic
features, (2) lexical features, and (3) Reddit user features. 
For the
psycholinguistic features, in line with much previous work, we used LIWC
\citep{pennebaker2015development}, a widely used tool in predicting mental
health, which classifies the words into dictionary-defined categories.  We
extracted 93 features, including syntactic features (e.g., pronouns, articles),
topical features (e.g., work, friends), and psychological features (e.g.,
emotions, social context).  In addition to LIWC, we used Empath
\citep{fast2016empath}, which is similar to LIWC but categorizes the words
using similarities based on neural embeddings. We used the 200 predefined and
manually curated categories, which \citeauthor{fast2016empath}~have found to be
highly correlated with LIWC categories ($r$=0.906).

The lexical features are the tf-idf weighted bag-of-words, stemmed using Porter
stemmer from NLTK \cite{bird2009natural}. Finally, Reddit user features are meant to
model user's interaction patterns. These include post-comment ratio,
the number of \emph{gilded} posts (posts awarded with money by other users),
average controversiality, the average difference between \emph{ups} and
\emph{downs} on user's comments and the time intervals between comments (the
mean, median, selected percentiles, and the mode).\footnote{Users with bipolar
disorder often experience sleep disturbance, which can make their usage 
patterns deviate from that of other users. Unfortunately, timestamps in Big
Query are in UTC, not in users' local times, thus determining the time zone
would require geolocalization. We leave this for future work.}

\paragraph{Experimental setup.}

We frame bipolar disorder prediction as a binary classification task, using the above-defined features and three classifiers: a support vector machine (SVM), logistic regression, and random forest ensemble (RF).
We evaluated our models and tune the hyperparameters using 10$\times$5 nested cross validation. %, with $k = 10$ in outer and $k = 5$ in inner loop.
%Feature selection is done by selecting $N$ features with the lowest p-value after the t-test, with $N$ ranging from $50$ to $5000$.
To mitigate for class imbalance, we use class weighting when training classifiers on the dataset split into categories.
As baselines, we used a majority class classifier (MCC) for evaluating the accuracy score and a random classifier with class priors estimated from the training set for evaluating the F1-score (F1-score is undefined for MCC).
For implementation, we used Scikit-learn \citep{pedregosa2011scikit}. We use a two-sided t-test for all statistical significance tests and test at p$<$0.001 level.

\paragraph{Results.}

\begin{table}[t]
\centering
{\small
\begin{tabular}{lrr}
\toprule
\multicolumn{1}{l}{} & \multicolumn{1}{c}{Acc} & \multicolumn{1}{c}{F1} \\
\midrule
MCC                   & 0.529                       &  --                     \\
Random                & 0.546                       & 0.453                  \\
SVM                   & 0.865                       & 0.853                  \\
%LinSVM                & 0.864                       & 0.852                  \\
LogReg                & 0.866                       & 0.849                  \\
RF                    & \textbf{0.869}              & \textbf{0.863}        \\
\bottomrule
\end{tabular}}
\caption{\label{tbl:rez_all} Prediction accuracy and F1-scores}
\end{table}

\begin{table}[t]
\centering
{\small
\begin{tabular}{lrrrr}
\toprule
       & LIWC  & Empath & Tf-idf & All   \\ 
\midrule
SVM    & 0.837 & 0.782  & \textbf{0.865}  & 0.838 \\
%LinSVM & 0.833 & 0.818  & \textbf{0.864}  & 0.835 \\
LogReg & 0.841 & 0.819  & \textbf{0.866}  & 0.862 \\
RF     & 0.829 & 0.825  & 0.869  & \textbf{0.869} \\
\bottomrule
\end{tabular}}
\caption{\label{tbl:rez_features} Prediction accuracy for the different models and feature sets}
\end{table}

\begin{table}[t]
\centering
{\small
\begin{tabular}{lrr}
\toprule
                      & MCC   & Our models \\
\midrule
Animals               & 0.693 & 0.807*      \\
AskReddit             & 0.606  & 0.856*      \\
Gaming                & 0.754  & 0.777*     \\
Jobs and finance      & 0.665 & 0.752*      \\
Movies/music/books    & 0.761 & 0.817*      \\
Politics              & 0.880  & 0.882*      \\
Religion              & 0.724  & 0.784*      \\
Sex and relationships & 0.513 & 0.801*      \\
Sports                & 0.832  & 0.837\phantom{*}     \\
\bottomrule
\end{tabular}}
\caption{\label{tbl:rez_cat} Accuracy of the MCC baseline and our models across topic categories.
Accuracies marked with ``*'' are significantly different from the baseline.}
\end{table}

Table \ref{tbl:rez_all} shows the accuracy and F1-scores for the different classifiers.
Random forest classifier achieved the best results, with accuracy of 0.869 and
an F1-score of 0.863.  All models outperform the baseline accuracies of 0.529 and 0.546, and the baseline F1-score of 0.453.

Table \ref{tbl:rez_features} shows the accuracy of the models using different
feature sets.  We observe two trends: Empath generally performs worse than
LIWC, and tf-idf features perform better than LIWC. However, looking at
the scores of the random forest classifier as the best model, we find that there is no significant difference between LIWC and Empath. 
Tf-idf does perform significantly different than both LIWC and Empath, while all
features combined (including Reddit user features) do not differ from tf-idf
alone. We speculate that tf-idf might yield better results in
this case because essentially all the words that LIWC and Empath detect also
exist as individual features in tf-idf. Similarly,
\citet{coppersmith2014quantifying} achieve better results using language models
than LIWC, arguing that many relevant text signals go undetected by
LIWC.

Finally, Table \ref{tbl:rez_cat} shows the accuracy across topic categories for
the MCC baseline and the best classifier in each category. 
Our models outperform MCC in all categories, and the differences are significant for all categories 
except \emph{Sports}. 
%\alert{Here we use MCC as baseline, instead of
%random classifier, to have a higher mountain to climb.}\todo{not sure what you
%mean. MCC goes with acc and Random with F1, both are competitive | nije mi bas
%neka recenica da...ali MCC je uvijek bio jaci od Randoma, pa smo ga zato uzeli.
%jer je imalanced dataset}

%because essentially, both of those tools are simple word counters, just like
%the bag-of-words approach.  The difference is that bag-of-words counts every
%word it comes across of, while Empath and LIWC look for that word in limited
%dictionaries.  

%\begin{table}[]
%\begin{tabular}{l|lllll}
%       & LIWC  & Empath & Features & Tf-idf & All   \\ \hline
%SVM    & 0.837 & 0.782  & 0.651    & 0.865  & 0.838 \\
%LinSVM & 0.833 & 0.818  & 0.552    & 0.864  & 0.835 \\
%LogReg & 0.841 & 0.819  & 0.732    & 0.866  & 0.862 \\
%RF     & 0.829 & 0.825  & 0.799    & 0.869  & 0.869
%\end{tabular}
%\caption{\label{tbl:rez_features} Accuracy for different feature sets, on the complete dataset.}
%\end{table}

%% file: feature.tex
\section{Feature Analysis}

We analyze the merit of the psycholinguistic features using a two-sided t-test,
with the null hypothesis of no difference in feature values between users with
bipolar disorder and control users. The lower the p-value, the higher the merit. We
analyzed the features separately on the entire dataset and on the 
dataset split into categories.

\paragraph{Between-group analysis.}

Ten LIWC features with the lowest p-value on the entire dataset are presented
in Table~\ref{tbl:liwc}, together with feature value means for the two groups. The values in the table are percentages of words in text from each category, except \emph{Authentic} and \emph{Clout}, which
are ``summary variables'' devised by \citet{pennebaker2015development}.
Personal pronouns, especially the pronoun \emph{I}, are used more often by
bipolar disorder users. This observation is in accord with past studies on
language of depressed people, which we can compare to because a bipolar
depressive episode is almost identical to major depression
\citep{anderson2012bipolar}.  \citet{coppersmith2014quantifying} also report a
significant difference in the use of \emph{I} between Twitter users with
bipolar disorder and the control group. 
The \emph{Authentic} feature of \citet{newman2003lying} reflects the authenticity of the author's text: a
higher value of this feature in bipolar disorder users may perhaps be
explained by them speaking about personal issues more sincerely, though further
research would be required to confirm this. 
We also observe a higher use of
words associated with feelings (\emph{feel}), \emph{health}, and biological
processes (\emph{bio}).  \citet{kacewicz2014pronoun} argue that pronoun use
reflects standings in social hierarchies, expressed through \emph{Clout} and
\emph{power} features: we observe a lower use of these words in users with
bipolar disorder, which might indicate they think of themselves as
less valuable members of society.  
The analysis of Empath features yielded
similar findings: \emph{health}, \emph{contentment}, \emph{affection},
\emph{pain}, and \emph{nervousness} have higher values in users 
with bipolar disorder.

\paragraph{Per-category analysis.} Significant features in specific categories
follow a pattern similar to the features on the complete dataset.  Pronoun
\emph{I} is statistically significant in all of the categories, as well as
features \emph{Clout} and \emph{Authentic}.  

\begin{table}[]
\centering
{\small
\begin{tabular}{lrr}
\toprule
Feature    & bipolar $\mu$    & control $\mu$    \\ 
\midrule
Authentic        & 52.65 & 32.92 \\
ppron             & 10.69 & 8.66  \\
i                & 5.84  & 3.38  \\
health           & 0.96  & 0.50  \\
feel       & 0.69  & 0.48  \\
power     & 2.11  & 2.58  \\
pronoun    & 16.87 & 14.86 \\
bio         & 2.65  & 1.90  \\
Clout       & 48.51 & 58.03 \\
article     & 5.88  & 6.55 \\
%conj   & 6.70 &  6.04 \\
%Analytic  & 41.27 & 50.82 \\
%they   &  0.88 &   1.11 \\
%anger   &  0.81 &  1.12 \\
%time   & 4.60 &  4.06 \\
%Tone & 52.16 & 41.92 \\    % maybe interesting? Emotional tone
%posemo & 3.89 &   3.44 \\
%sad & 0.45 &  0.36 \\
\bottomrule
\end{tabular}}
\caption{\label{tbl:liwc} Mean values of most significant LIWC features for both groups}
\end{table}

%As mentioned, \citet{rude2004language} find the excessive use of negatively valanced words in language of depressed students, and slightly lesser use of positive words ($p=0.8$).
%We observe higher use of words associated with sadness ($\mu_b = 0.46$, $\mu_c = 0.36$) and anxiety ($\mu_b = 0.36$, $\mu_c = 0.27$) in bipolar group, but higher use of anger words ($\mu_b = 0.81$, $\mu_c = 1.12$) in the control group. % resulting in higher use of negative words in control group
%The excessive use of sadness and anxiety words can be expected, and is inline with previous research about depression.
%However, we also find the bipolar authors use significantly more positive emotion words ($\mu_b = 3.89$, $\mu_c = 3.44$) than the control group.
%This can be due to the characteristics of manic episodes, which do not occur in clinically depressed people.
%The observations are the same in features generated with LIWC and with Empath, with reported means being from LIWC.

%% file: emotions.tex
\section{Emotion Analysis}

%\citet{coppersmith2014quantifying} report that LIWC emotion-expressive features
%correlate with author's mental state \todo{"ovo mi malo smrdi" -- what's the
%problem exactly? | pa zapravo nema problema...mislim da je malo i zalutala recenica.
%Htio sam nekako povezati zasto mislimo da bi LIWC trebao ukazati na nesto, ali
%mozda nepotrebno? U drugom paperu sam nasao da LIWC korelira sa stvarnim emocijama,
%citirajuci ovaj coppersmith2014, ali u tom originalu nema nesto takvo receno.}.
As emotional swings are of the main symptoms of bipolar
disorder, we expect that there will be a difference in the use of
emotion words between users with bipolar disorder and the control
group. We report the results for LIWC, as Empath gave very similar results.

%\begin{figure}
%    \centering
%    \hspace*{-1em}\includegraphics[scale=0.25]{nova_figura.png}
%    \caption{Boxplot of values of LIWC emotion categories for bipolar and control groups}
%    \label{fig:emo1}
%\end{figure}

\begin{table}
\centering
{\small
\begin{tabular}{lrr}
\toprule
& Bipolar & Control \\
\midrule
posemo & 3.899 $\pm$ 1.02 & 3.442 $\pm$ 0.78 \\
negemo & 2.432 $\pm$ 0.67 & 2.569 $\pm$ 0.70 \\
anxiety & 0.367 $\pm$ 0.19 & 0.266 $\pm$ 0.10 \\
anger & 0.818 $\pm$ 0.39 & 1.128 $\pm$ 0.52 \\
sad & 0.455 $\pm$ 0.21 & 0.363 $\pm$ 0.11 \\
affect & 6.415 $\pm$ 1.22 & 6.074 $\pm$ 1.12 \\
\bottomrule
\end{tabular}}
\caption{\label{tbl:emoemo} Means and standard deviations of LIWC emotion categories for bipolar and control group} 
\end{table}

\paragraph{Between-group differences.}

Table \ref{tbl:emoemo} shows means and standard deviations of the values of six
LIWC emotion categories (\emph{posemo}, \emph{negemo}, \emph{anxiety},
\emph{anger}, \emph{sad}, and \emph{affect}) for the users with bipolar
disorder and the control group. Users with bipolar disorder use significantly
more words linked with general affect.  Furthermore, we observe increased use
of words related to sadness, while the control group uses more anger-related
words.  The results for \emph{sadness} are in line with previous work on
depressed authors. In addition, we find significant use of \emph{anxiety} words
in users with bipolar disorder, similar to the findings of
\citet{coppersmith2014quantifying}. Surprisingly, we find that users with
bipolar disorder use more positive emotion words than the control group. This
is in contrast to findings of \citet{rude2004language}, who report no
statistical significance in the use of positive emotion words in
depressed authors. We speculate that this difference may be due to the
characteristics of manic episodes, which do not occur in clinically depressed
people. 

\paragraph{Per-category differences.}
 
The difference between users with bipolar disorders and the control group in
\emph{AskReddit}, \emph{Animals}, \emph{Movies/music/books}, and \emph{Sex and
relationships} categories is significant in words related to
sadness, anxiety, anger, and positive emotions. However, there is no
significant difference in positive and negative emotions in categories
\emph{Jobs} and \emph{Politics}, while \emph{Sports}, \emph{Gaming}, and
\emph{Religion} differ only in positive emotions.

\paragraph{User-level variance.}

We hypothesize that, due to the alternation of manic and depressive episodes,
users with bipolar disorder will have a higher variance across time in the use
of emotion words than users from control group. To verify this, we randomly
sampled 100 users with bipolar disorder and 100 control users from all the
users in our dataset with more than 100K words and split their comments into
monthly chunks. For each of the 200 users, we calculated the LIWC features for each month
and computed their standard deviations. We then measured the difference between standard deviations for the two groups.
Table~\ref{tbl:emo} shows the results. We find that bipolar users have
significantly more variance in most emotion-expressive words, which
confirms our hypothesis.

\begin{table}
\centering
{\small
\begin{tabular}{lrrr}
\toprule
        & Bipolar       & Control  & p-value     \\
\midrule
posemo  & 0.00272* & 0.00166 & 0.00272\\
negemo  & 0.00583*  & 0.00379 & 0.00583\\
anxiety & 0.00765* & 0.00627 & 0.00765\\
anger   & 0.01745\phantom{*}  & 0.01422 & 0.01745 \\
sadness & 0.00695* & 0.00572 & 0.00695\\
\bottomrule
\end{tabular}}
\caption{\label{tbl:emo} Averages of standard deviations in the use of emotion-expressive words for the two groups. All differences are significant except for ``anger''.} 
\end{table}

%% file: discussion.tex
\section{Conclusion}

We presented a preliminary study on bipolar disorder prediction from 
user comments on Reddit.  Our classifiers outperform
the baselines and reach accuracy and F1-scores of above 86\%.  Feature
analysis suggests that users with bipolar disorder use more first-person
pronouns and words associated with feelings. They also 
use more affective words, words related to sadness and anxiety, but
also more positive words, which may be explained by the alternating episodes.
There is also a higher variance in emotion words across time in
users with bipolar disorder.
Future work might look into the linguistic differences in manic
and depressive episodes, and propose models for predicting them.  

%Also, deep
%learning is to be used for better classification.